\documentclass{article}
\usepackage{arxiv}
\usepackage[T1]{fontenc}
\usepackage[utf8]{inputenc}
\usepackage{listings}
\usepackage{booktabs}
\usepackage{multirow}
\usepackage{xurl}
\usepackage{graphicx}
\usepackage{natbib}

\title{Adaptations of AI models for querying the LandMatrix database in natural language}

\usepackage{authblk}

\author[1,2]{Fatiha Ait Kbir \texttt{aitkbirfati@gmail.com}}
\author[3,4,5]{Jérémy Bourgoin  \texttt{prenom.nom@cirad.fr}}
\author[1,4]{Rémy Decoupes \texttt{prenom.nom@inrae.fr}}
\author[5]{Marie Gradeler \texttt{m.gradeler@ifad.org}}
\author[3,4]{Roberto Interdonato \texttt{prenom.nom@cirad.fr}}

\affil[1]{INRAE, F-34398 Montpellier, France}
\affil[2]{Intern at CIRAD, F-34398 Montpellier, France}
\affil[3]{CIRAD, F-34398 Montpellier, France}
\affil[4]{TETIS, Univ. Montpellier, AgroParisTech, CIRAD, CNRS, INRAE, Montpellier 34090, France}
\affil[5]{International Land Coalition, Rome, Italy}

\begin{document}
\maketitle
\begin{abstract}
The Land Matrix initiative (https://landmatrix.org) and its global observatory aim to provide reliable data on large-scale land acquisitions to inform debates and actions in sectors such as agriculture, extraction, or energy in low- and middle-income countries. Although these data are recognized in the academic world, they remain underutilized in public policy, mainly due to the complexity of access and exploitation, which requires technical expertise and a good understanding of the database schema.

The objective of this work is to simplify access to data from different database systems. The methods proposed in this article are evaluated using data from the Land Matrix. This work presents various comparisons of Large Language Models (LLMs) as well as combinations of LLM adaptations (Prompt Engineering, RAG, Agents) to query different database systems (GraphQL and REST queries). The experiments are reproducible, and a demonstration is available online: https://github.com/tetis-nlp/landmatrix-graphql-python.
\end{abstract}



\section{Introduction}

The Land Matrix Initiative (LMI) is an independent land monitoring initiative that collates and seeks to verify information on large-scale land acquisitions. It records transactions that entail a transfer of rights to use, control or own land through sale, lease or concession that cover 200 hectares (ha) or larger, and that have been concluded since the year 2000 \citep{anseeuw_transnational_2012}.

LMI database provides two Application Programming Interfaces (APIs), REST and GraphQL, which can be used to retrieve data by sending specific requests. For example, to identify deals larger than 1,000 hectares that have been canceled since 2016, the corresponding REST request would be: \url{https://landmatrix.org/api/deals/?area\_min=1000\&negotiation\_status=CONTRACT\_CANCELED\&initiation\_year\_min=2016}

Despite being a global reference on land acquisition phenomena in the academic community, the LMI’s data remains underutilized in supporting public action, even with considerable efforts to improve accessibility. Effectively accessing and utilizing this data presents significant challenges, as it requires a deep understanding of both the dataset and query interfaces to extract relevant and actionable insights for the various stakeholders involved. 

The primary objective is to simplify access to the LMI database. To achieve this, we adapted the approaches used in Text-to-SQL to adjust LLMs for generating REST and GraphQL requests. This adaptation allows users to interact with the database more easily by enabling LLMs to produce executable queries that retrieve the desired information.
Text-to-SQL is the process of translating natural language questions into database-executable SQL queries. Given a user question, the LLM takes the question and its corresponding database schema as input and then generates an SQL query as output. This query is then executed to retrieve the information needed to answer the user's question \citep{hong_next-generation_2024}.

Early research efforts in Text-to-SQL focused on fine-tuning Transformer models or decoding techniques by integrating SQL syntax, semantics, and the intricate relationship between questions and databases. The emphasis was largely on using prompting strategies to transform user queries into SQL statements. Recently, more advanced prompting techniques have emerged, improving the interpretation of natural language questions and database schemas. These innovations include selecting relevant few-shot examples based on query similarity, breaking down complex questions into smaller tasks, verifying the accuracy of generated SQL through execution feedback, and linking natural language phrases to corresponding database elements \citep{sun_sql-palm_2024}.

While these few-shot prompting methods have significantly enhanced Text-to-SQL performance, it remains uncertain whether prompting alone is sufficient to address the challenges posed by the LMI database. This is especially true when user queries are unclear or highly complex, which can hinder the LLM’s ability to identify and process all relevant elements in the question. For instance, if a user is searching for deals related to a specific country, the model may occasionally fail to correctly identify the country and link it to its corresponding ID.

The contribution of this article is to compare different LLMs with various optimizations for extracting data from the Land Matrix using natural language questions. The study evaluates three open-weight models, Llama3-8B, Mixtral-8x7B-instruct, and Codestral-22B using a dataset of real non-technical user requests in natural language and their expert-configured REST and GraphQL queries. The work is reproducible and allows for the inclusion of additional models. Three optimization techniques are compared: prompt engineering, retrieval-augmented generation, and LLM agents.

\section{Related work}
The generative capabilities of language models offer new opportunities to interact with database systems without necessarily needing to know the syntax. Since 2017 and the publication of "Attention is All You Need" \citep{vaswani_attention_2017}, natural language processing has seen a rapid improvement in performance thanks to transformers, a new neural network architecture. This layer allows language models to better grasp the mechanisms of languages. After the phase of encoder-based models, such as BERT \citep{devlin_bert_2019}, which are non-generative but enable handling various classic NLP tasks (named entity recognition, text classification, sentiment analysis, etc.), larger models with generative capabilities have become popular, particularly with the release of GPT-3.5 by OpenAI through its ChatGPT service.

These personal assistants have various capabilities for generating text, such as answering questions, summarizing, and translating into other languages \citep{yang_harnessing_2023, zhao_survey_2023}. They can also generate computer languages, such as programming languages \citep{roziere_code_2024} or database query languages \citep{liu_suql_2024}. Moreover, according to the study by \cite{zheng_lmsys-chat-1m_2023}, the topics most frequently addressed by chatbots are related to code generation (debugging, solving errors, code generation).

To achieve better results and more accurate database query requests, several techniques can be used. The first, called prompt engineering \citep{brown_language_2020}, involves carefully crafting the questions addressed to the chatbot to obtain the best possible results. The structure of the prompt varies depending on the model and its fine-tuning, so a good understanding of the model is necessary. The second option, few-shot engineering \citep{brown_language_2020}, involves providing a series of examples—four or five—within the prompt to help the model respond more effectively. Another approach is to feed the prompt with external knowledge via an up-to-date database. Known as RAG, Retrieval Augmented Generation, \citep{lewis_retrieval-augmented_2021}, this method provides relevant information to assist the model in formulating its query (such as a list of possible column values). Other methods include retraining the model through fine-tuning or alignment \citep{balaguer_rag_2024}, but these require significant computational resources.

\section{Methodology}
The objective of this work is to compare different LLMs with various optimizations to extract data from the Land Matrix using natural language questions. To achieve this, different Land Matrix partners gathered a set of almost 60 REST or GraphQL query requests from real non-technical users, and then added the queries they themselves configured.

In this study, we compare three open-weight models: Llama3-8B, a Mixture of Experts model aligned on an instruction dataset (Mixtral-8x7B-instruct), and a code-generation-specialized LLM (Codestral-22B). This work is reproducible, and additional models can be included in the comparison as long as APIs of the OpenWebUI type are available \footnote{https://github.com/open-webui/open-webui}.

Finally, we propose three optimizations. The first is prompt engineering, the second is RAG and the last one involves LLM Agents.

\subsection{Prompt Engineering:}
\begin{figure}[ht]
    \centering
    \includegraphics[width=0.9\textwidth]{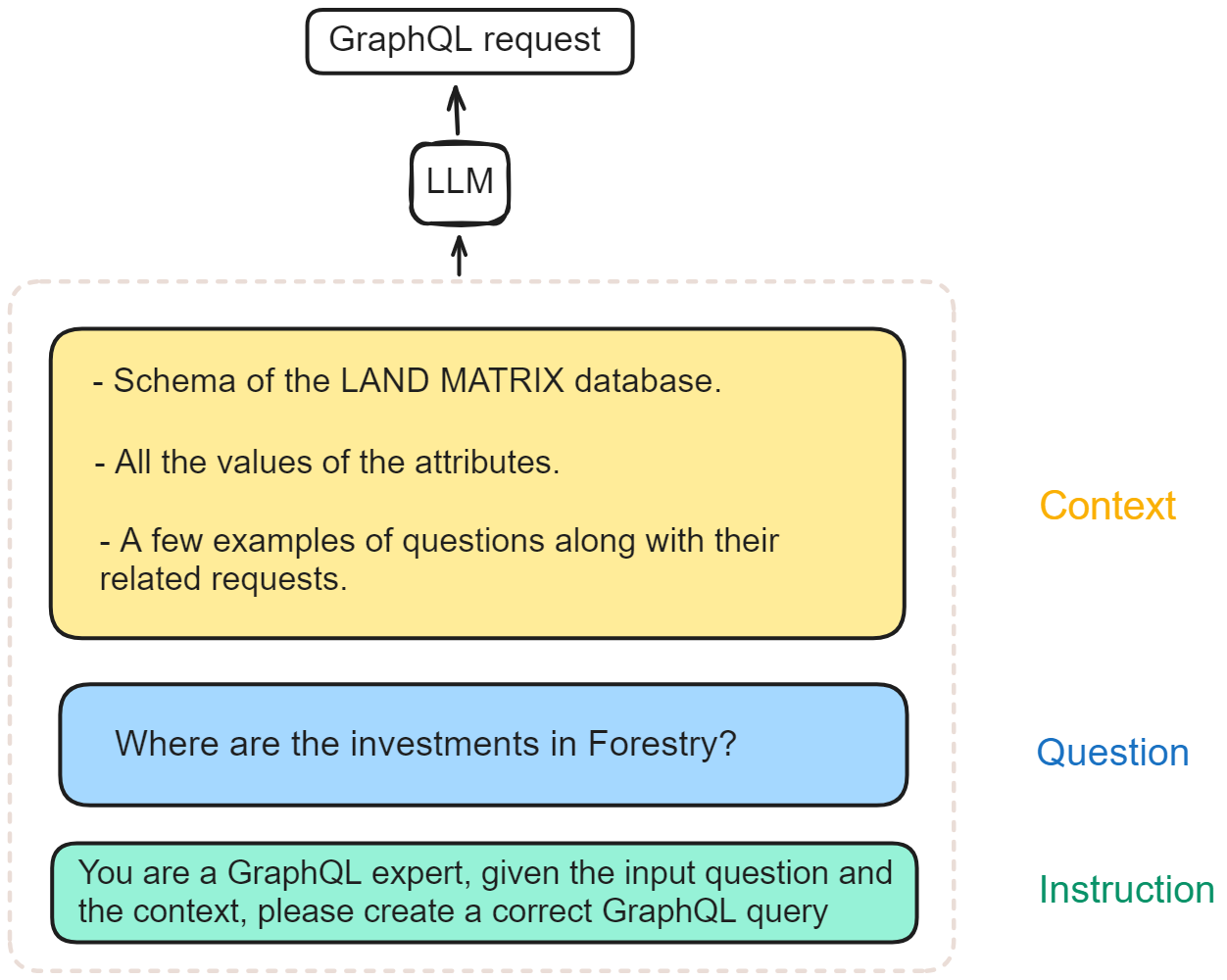}
    \caption{Detailed prompt used with three sections: role given to the LLM (instruction), an example of real natural language question (question), and a context with few-shot learning with the database schema (context).}
    \label{fig:PromptEng}
\end{figure}

We employed a prompt specifically designed to generate (GraphQL | REST) requests. As illustrated by Fig.\ref{fig:PromptEng}, This prompt takes into account the following:

\begin{itemize}
    \item The user's question as input.
    \item Context that includes the database schema, detailing the attributes and their relationships, as well as a list of all possible attributes and their corresponding values in the LMI database, examples of question-request pairs to provide the model with an understanding of request syntax. Additionally, we incorporated rules and information regarding the LMI API.
    \item We instructed the LLM to act as a (GraphQL | REST) specialist, capable of generating meaningful queries in response to the initial question.
\end{itemize}

\subsection{Retrieval augmented generation (RAG):}

To enhance the quality of the GraphQL and REST queries generated by the model, we employed the RAG approach to enrich the prompt context with natural language questions and their corresponding queries that are similar to the input question.
The process begins by calculating the embeddings of the input question and the natural language questions in our question-query corpus using the all-mpnet-base-v2 model. This sentence-transformer model maps sentences and paragraphs into a dense 768-dimensional vector space, which is useful for tasks such as clustering or semantic search. Next, we utilized Facebook AI Similarity Search (Faiss), a library for similarity search and clustering of dense vectors. Faiss includes algorithms capable of searching within vector sets of any size, even those that don't fit in RAM, and provides support for evaluation and parameter tuning.
We calculate the similarity between the input question and all the questions represented by their embeddings stored in the vector database. We then return the top k most similar questions to the input question, along with their corresponding (GraphQL | REST) queries.
Instead of using random queries in this context, this method enriches the context with queries similar to the input question.

\subsection{Agents:}

\begin{figure}[ht]
    \centering
    \includegraphics[width=1\textwidth]{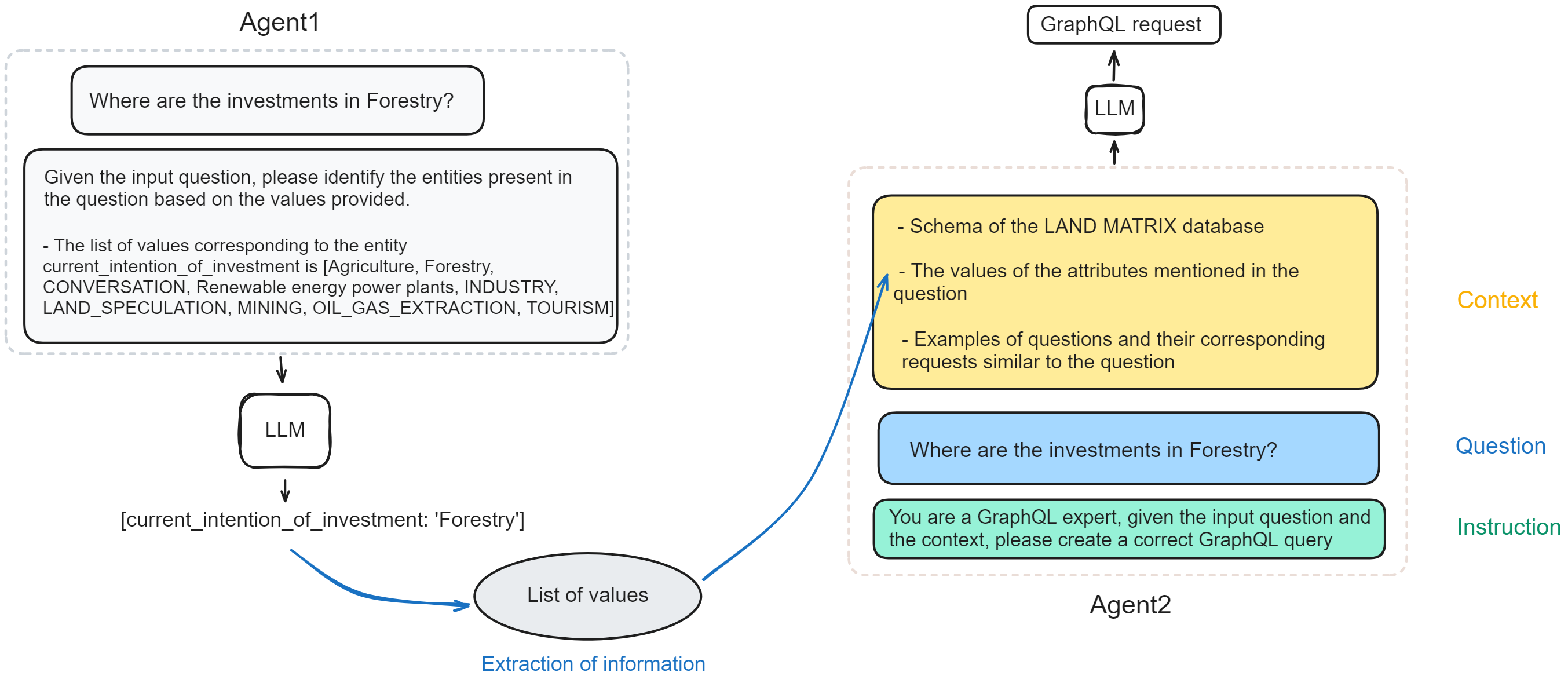}
    \caption{Pipeline of the Agents optimization: another language model is used to extract the filters wanted by the user and its outputs are added to the previous prompt.}
    \label{fig:Methods}
\end{figure}

Multi-Agent Collaboration involves combining multiple AI agents to divide tasks and exchange information in order to optimize performance. In our case, we utilized two agents: the first for entity extraction and the second for query generation (see Fig. 2).

We start by extracting the entities from the user’s question using the first agent, where predefined values associated with entities in the LMI database are returned as a list of [entity: value] pairs. This list is then used to retrieve all relevant values for these attributes from a predefined set, enriching the context for the second agent. The second agent is responsible for generating the query by incorporating the relevant information about the attributes mentioned in the question. Additionally, we enhanced the process by including similar question-query pairs using the RAG method and relevant database information.

This approach allows us to enrich the query generation prompt with context-specific information without overloading it with all possible variables. Overloading the prompt can reduce the model's ability to accurately process the information, resulting in suboptimal performance.


\section{Experiments}

The experiments aim to assess the effectiveness of LLMs with the three optimizations. To do this, we evaluate the generated queries from three perspectives. The first consists of verifying that the syntax of the generated query is correct, meaning that it does not produce an error when submitted to the Land Matrix API. The second measures the similarity between the generated query and the one manually created by the Land Matrix database administrators. Finally, the last evaluation checks whether the data retrieved by the generated query accurately covers the data obtained by the manually created query.

\textbf{Data:}
LMI provides APIs for accessing data. By executing (GraphQL | REST) queries, the results are returned in JSON format. To adapt our LLM for generating these queries, we prepared a corpus of natural language questions along with their corresponding queries. We divided the corpus into two parts: one to fine-tune the LLM on the structure of (GraphQL | REST) queries, and the other to test the model's performance and its ability to generate the queries correctly.

\subsection{Evaluation metrics:}
To evaluate the validity of the queries generated by our LLMs, we identified specific metrics for assessment. These include metrics for evaluating the overall quality of the queries and those for assessing the attributes used within them like. The goal is to determine whether the model accurately identified the correct attributes and values necessary for effective search operations.

\subsubsection{Metrics for global queries}
 \textbf{Jaccard Similarity} is a metric used to evaluate the similarity and diversity between sets. 
 
\[ J(A,B) = \frac{|A \cap B|}{|A \cup B|} \]

We used Jaccard similarity to calculate the similarity between the results of the queries generated by the model and the actual queries in our corpus. To evaluate the queries across the entire corpus, we calculated the average of these similarity values.

\begin{itemize}
    \item \textbf{Valid Query} is a metric used to assess the syntactic correctness of generated queries by calculating the Jaccard Similarity between the real queries and the generated queries.
\end{itemize}

\begin{itemize}
    \item \textbf{Valid Result} is a metric used to evaluate the similarity between real query results and generated query results using Jaccard Similarity.
\end{itemize}

\subsubsection{Metrics for attributes}

To evaluate the attributes, we calculated the following metrics: precision, recall, and F1-score. For this, we identified true positives (TP), false positives (FP), and false negatives (FN) as follows:

\textbf{TP}: cases where the relevant attributes are found in both queries, indicating the model has correctly identified the necessary attributes.

\textbf{FP}: cases where the model generated additional elements beyond the required attributes, signaling overproduction or inclusion of irrelevant information.

\textbf{FN}: cases where the model failed to identify the relevant attributes, meaning it missed essential elements in its response.

\textbf{TP-values}: refers to the validity of the values assigned to the attributes.

\section{Results}

The results presented in this section are divided into three parts: verification of the syntax of the generated queries, the similarity between the generated query and the ground truth, and the coverage of the data obtained through the generated query compared to the ground truth for both LMI APIs, GraphQL and REST.

\subsection{Validity of REST and GraphQL request syntax}
Table \ref{tab:syntaxe_validity} presents the results concerning the validity of the REST or GraphQL generated requests. Llama3 and Mixtral struggle to generate valid queries for the REST API, only Codestral is capable, with agents, to generate REST queries. 
Concerning GrapQL, it seems that LLMs are more comfortable with this syntax. We can notice as well that agentic method is the best approach and prompt-engineering performs better than RAG. 

\begin{table}[ht]
    \centering
    \begin{tabular}{llcc}  
        \toprule
        \textbf{Approach} & \textbf{LLM} & \textbf{Valid Query for REST} & \textbf{Valid Query for GraphQL}\\
        \midrule
        \multirow{3}{*}{Prompt-Engineering} & Llama3 & 0\% & 65\% \\
                                    & Mixtral 8x7B & 0\% & 42\% \\
                                    & Codestral-22B & 2\% & 69\% \\
        \midrule
        \multirow{3}{*}{RAG} & Llama3 & 0\% & 49\% \\
                                    & Mixtral 8x7B & 0\% & 44\% \\
                                    & Codestral-22B & 29\% & 56\% \\
        \midrule
        \multirow{3}{*}{Agentic} & Llama3 & 2\% & 73\% \\
                                    & Mixtral 8x7B & 0\% & 72\% \\
                                    & Codestral-22B & \textbf{52}\% & \textbf{77}\%\\
        \bottomrule
    \end{tabular}
    \caption{Validity of the syntax of the requests generated based on the approaches and models.}
    \label{tab:syntaxe_validity}
\end{table}

\subsection{Accuracy of filters in requests}

This evaluation considers only syntactically valid queries. As observed in Table \ref{tab:syntaxe_validity}, some methods generate few or no valid queries. We evaluate the ability of these methods to produce queries similar to the ground truth based on these valid queries, by comparing the list of filters between the generated queries and the queries from the dataset.
Table \ref{tab:filter_accuracy_rest} shows the results for REST queries whereas Table \ref{tab:filter_accuracy_graphql} presents scores for GraphQL API. By comparing those two tables, we can notice that even if it is harder to generate syntactically correct REST queries than GraphQL queries, models achieve better performance on valid REST requests.

\begin{table}[ht]
    \centering
    \begin{tabular}{llccccc}  
        \toprule
        \textbf{Approach} & \textbf{LLM} & \textbf{Precision} & \textbf{recallL} & \textbf{accuracy} & \textbf{f1-score} & \textbf{value-score}    \\
        \midrule
        \multirow{3}{*}{Prompt-Engineering} & Llama3 & -& -& -& -& -\\
                                    & Mixtral 8x7B & -& -& -& -& -\\
                                    & Codestral-22B & 100\% & 100\% & 100\%& 100\% & 2\%  \\
        \midrule
        \multirow{3}{*}{RAG} & Llama3 & -& -& -& -& -\\
                                    & Mixtral 8x7B & -& -& -& -& -\\
                                    & Codestral-22B & 67\% & 61\% & 46\% & 64\% & 40\%\\
        \midrule
        \multirow{3}{*}{Agentic} & Llama3 & 100\% & 36\% & 36\%& 53\% & 2\%\\
                                    & Mixtral 8x7B & - & - &  - & - & - \\
                                    & Codestral-22B & 88\% & 74\% & 67\%& 80\% & 59\%\\
        \bottomrule
    \end{tabular}
    \caption{Accuracy of filters in REST requests for syntactically valid queries}
    \label{tab:filter_accuracy_rest}
\end{table}

\begin{table}[ht]
    \centering
    \begin{tabular}{llccccc}  
        \toprule
        \textbf{Approach} & \textbf{LLM} & \textbf{Precision} & \textbf{recallL} & \textbf{accuracy} & \textbf{f1-score} & \textbf{value-score}    \\
        \midrule
        \multirow{3}{*}{Prompt-Engineering} & Llama3 & 94\% & 55\% & 54\% & 70\% & 19\%  \\
                                    & Mixtral 8x7B & 92\% & 59\% & 40\% & 72\% & 12\% \\
                                    & Codestral-22B & 93\% & 82\% & 72\% & 87\% & 43\%  \\
        \midrule
        \multirow{3}{*}{RAG} & Llama3 & 75\% & 18\% & 16\% & 29\% & 5\%  \\
                                    & Mixtral 8x7B & 40\% & 13\% & 10\% & 19\% & 2\%\\
                                    & Codestral-22B & 79\% & 46\% & 38\% & 58\% & 23\%\\
        \midrule
        \multirow{3}{*}{Agentic} & Llama3 & 95\% & 71\% & 67\% & 81\% & 40\% \\
                                    & Mixtral 8x7B & 81\% & 71\% & 60\% & 76\% & 33\% \\
                                    & Codestral-22B & 88\% & 85\% & 76\% & 86\% & 48\%\\
        \bottomrule
    \end{tabular}
    \caption{Accuracy of filters in GraphQL requests for syntactically valid queries.}
    \label{tab:filter_accuracy_graphql}
\end{table}

\subsection{Accuracy of the data returned by the API}
The generated queries may differ slightly from the queries in the dataset but can still retrieve the expected set of data from the LMI API. Therefore, the Table \ref{tab:data_accuracy} presents the results of the data retrieved by the generated queries for the three models and their optimizations.

The observation of these results leads to the same conclusion: although the models struggle more with the REST API, their valid queries perform better than those for GraphQL. The best approach remains Codestral with the Agentic configuration.

\begin{table}[ht]
    \centering
    \begin{tabular}{llcc}  
        \toprule
        \textbf{Approach} & \textbf{LLM} & \textbf{Valid Result for REST} & \textbf{Valid Result for GraphQL} \\
        \midrule
        \multirow{3}{*}{Prompt-Engineering} & Llama3 & - & 15\% \\
                                    & Mixtral 8x7B & - & 2\% \\
                                    & Codestral-22B & 2\% & 48\% \\
        \midrule
        \multirow{3}{*}{RAG} & Llama3 & 1\% & 5\% \\
                                    & Mixtral 8x7B & 1\% & 3\% \\
                                    & Codestral-22B & 56\% & 30\% \\
        \midrule
        \multirow{3}{*}{Agentic} & Llama3 & 2\% & 37\% \\
                                    & Mixtral 8x7B & 2\% & 33\% \\
                                    & Codestral-22B & \textbf{66}\% & \textbf{51}\% \\
        \bottomrule
    \end{tabular}
    \caption{ Accuracy of the data returned by the requests generated based on the approaches and models.}
    \label{tab:data_accuracy}
\end{table}

\section{Discussion}
Based on the results, we can see that the Codestral-22B model performs well for both REST and GraphQL requests, maintaining a strong adherence to the requested format without any hallucinations or unnecessary comments, outperforming Llama3 and Mixtral 8x7B.  Codestral-22B captures the schema of the LMI database very accurately, as well as the relationships between its attributes, which is reflected in the higher accuracy scores in the filter tables.
LLaMA 3 also performs well compared to Mixtral, effectively capturing the database schema and attribute relationships. 
Mixtral, however, shows the lowest performance, as it is less specialized in programming languages and sometimes struggles with the complex schemas of LMI databases. Moreover, the version of Mixtral used in this article is an instruction fine-tuned version which leads the LLMs to not generate output when it is not enough confident.

For the methods, we observe that the performance of the models improves progressively from the in-context learning method to RAG, and finally to agents. This is due to the optimization of the prompt from using the entire database schema and related information, along with random examples of request responses, to tailoring the information specifically for each input question. This optimization makes the context easier for the model to process compared to handling a large, unstructured context. 

\section{Demonstration and reproducibility}
All the data used in this work and the codes developed for the experiments are available on our Github repository.

We also provide a small web application as a demonstrator. It is built on a Streamlit application incorporating the core components of a three-tier architecture: a client-side browser, web server application, and database. Additionally, the architecture features two key elements: an LLM-integration middleware, such as Langchain in our case, and a LLM. \cite{pedro_prompt_2023}

When a user submits a question, the chatbot calls the LangChain API, which collaborates with the LLM to interpret the query, generate a (GraphQL | REST) request, and retrieve the required information from the LMI database.

\section*{Conclusion}
The issue addressed in this work was whether the new code generation capabilities of models could improve the interface between a user and a database system. Based on data annotated by the Land Matrix database administrators, we propose an extensible comparison of three models with three adaptations (Prompt, RAG, and Agentic approach). The best combination, namely Codestral in Agentic mode, achieves two-thirds of the results expected by database users for REST and half for GraphQL.

Given that the Text-to-SQL task is extensively covered in the literature, we anticipate rapid improvements in models and approaches. For this reason, we have made our code available and ensured the experiments are reproducible, allowing other competitors to be added to our benchmark.

\section*{Acknowledgement}
This work has been realized with the support of the Land Matrix initiative and the ISDM-MESO Platform at the University of Montpellier funded under the CPER by the French Government, the Occitanie Region, the Metropole of Montpellier and the University of Montpellier.

\bibliographystyle{rnti}
\bibliography{biblio, biblio_remy}

\end{document}